\documentclass[10pt,twocolumn,letterpaper]{article}
\usepackage{iccv}
\usepackage{times}
\usepackage{epsfig}
\usepackage{graphicx}
\usepackage{amsmath}
\usepackage{amssymb}
\usepackage{booktabs}
\usepackage{array,multirow}
\usepackage{algorithm}
\usepackage{algorithmicx}
\usepackage[noend]{algpseudocode}
\usepackage{cuted}
\usepackage{capt-of}
\usepackage[marginal]{footmisc}
\usepackage[accsupp]{axessibility}

\usepackage[pagebackref=true,breaklinks=true,letterpaper=true,colorlinks,bookmarks=false]{hyperref}

\iccvfinalcopy 

\def\iccvPaperID{3801} 

\ificcvfinal\fi

\newcommand{\boldstart}[1]{\noindent\textbf{#1}}
\newcommand{\boldstartspace}[1]{\vspace{0.1in}\noindent\textbf{#1}}

\begin{document}

\title{MVSNeRF: Fast Generalizable Radiance Field Reconstruction\\ from Multi-View Stereo}

\author{Anpei Chen$^{*1}$
\and
Zexiang Xu$^{*2}$
\and
Fuqiang Zhao$^1$
\and
Xiaoshuai Zhang$^3$
\and
Fanbo Xiang$^{3}$
\and
Jingyi Yu$^1$ \qquad \qquad Hao Su$^3$\\
$^{1}$ ShanghaiTech University \qquad $^{2}$ Adobe Research \qquad $^{3}$ University of California, San Diego\\
{\tt\small {\{chenap,zhaofq,yujingyi\}}@shanghaitech.edu.cn \quad zexu@adobe.com \quad {\{xiz040,fxiang,haosu\}}@eng.ucsd.edu}
}



\maketitle
\ificcvfinal\thispagestyle{empty}\fi


\begin{strip}\centering
\includegraphics[width=\textwidth]{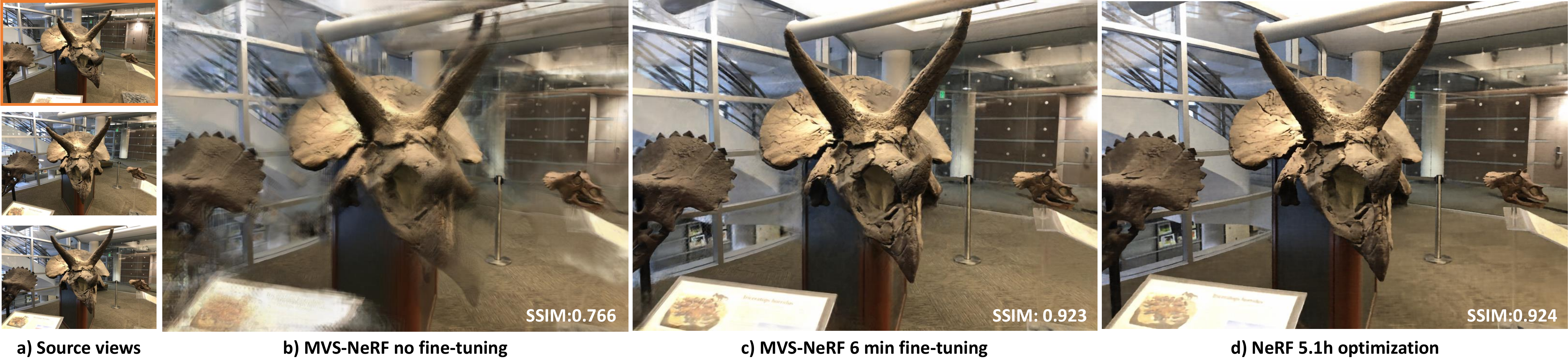}
\captionof{figure}{We train our MVSNeRF with scenes of objects in the DTU dataset \cite{dtu}. Our network can effectively \emph{generalize} across diverse scenes; even for a complex indoor scene, our network can reconstruct a neural radiance field from only three input images (a) and synthesize a realistic image from a novel viewpoint (b). While this result contains artifacts, it can be largely improved by fine-tuning our reconstruction on more images for only 6 min ($4k$ iterations) (c), which achieves comparable quality with the NeRF's \cite{nerf} result (d) from  5.1h per-scene optimization ($100k$ iterations). }
\label{fig:teaser}
\end{strip}

\newcommand{\hao}[1]{{\color{red}[hao: #1]}}
\footnote{$*$ Equal contribution}
\footnote{Research done when Anpei Chen was in a remote internship with UCSD.} 
\begin{abstract}
   We present MVSNeRF, a novel neural rendering approach that can efficiently reconstruct neural radiance fields for view synthesis.
   Unlike prior works on neural radiance fields that consider per-scene optimization on densely captured images,
   we propose a generic deep neural network that can reconstruct radiance fields from only three nearby input views via fast network inference.
   Our approach leverages plane-swept cost volumes (widely used in multi-view stereo) for geometry-aware scene reasoning, and combines this with physically based volume rendering for neural radiance field reconstruction.
   We train our network on real objects in the DTU dataset, and test it on three different datasets to evaluate its effectiveness and generalizability.
   Our approach can generalize across scenes (even indoor scenes, completely different from our training scenes of objects) and generate realistic view synthesis results using only three input images, significantly outperforming concurrent works on generalizable radiance field reconstruction.
   Moreover, if dense images are captured, our estimated radiance field representation can be easily fine-tuned; this leads to fast per-scene reconstruction with higher rendering quality and substantially less optimization time than NeRF.

\end{abstract}


\section{Introduction}
Novel view synthesis is a long-standing problem in computer vision and graphics.
Recently, neural rendering approaches have significantly advanced the progress in this area. Neural radiance fields (NeRF) and its following works~\cite{nerf,nerf_in_the_wild,Neural_sparse_voxel_fields} can already produce photo-realistic novel view synthesis results. 
However, one significant drawback of these prior works is that they require a very long per-scene optimization process to obtain high-quality radiance fields, which is expensive and highly limits the practicality.

Our goal is to make neural scene reconstruction and rendering more practical, by enabling \emph{highly efficient} radiance field estimation.
We propose MVSNeRF, a novel approach that \emph{generalizes well across scenes} for the task of reconstructing a radiance field from \emph{only several} (as few as three) unstructured multi-view input images. 
With strong generalizability, we avoid the tedious per-scene optimization and can directly regress realistic images at novel viewpoints via fast network inference. If further optimized on more images with only a short period (5-15 min), our reconstructed radiance fields can even outperform NeRFs \cite{nerf} with hours of optimization (see Fig.~\ref{fig:teaser}). 

\begin{figure*}[t]
\begin{center}
    \includegraphics[width=0.95\linewidth]{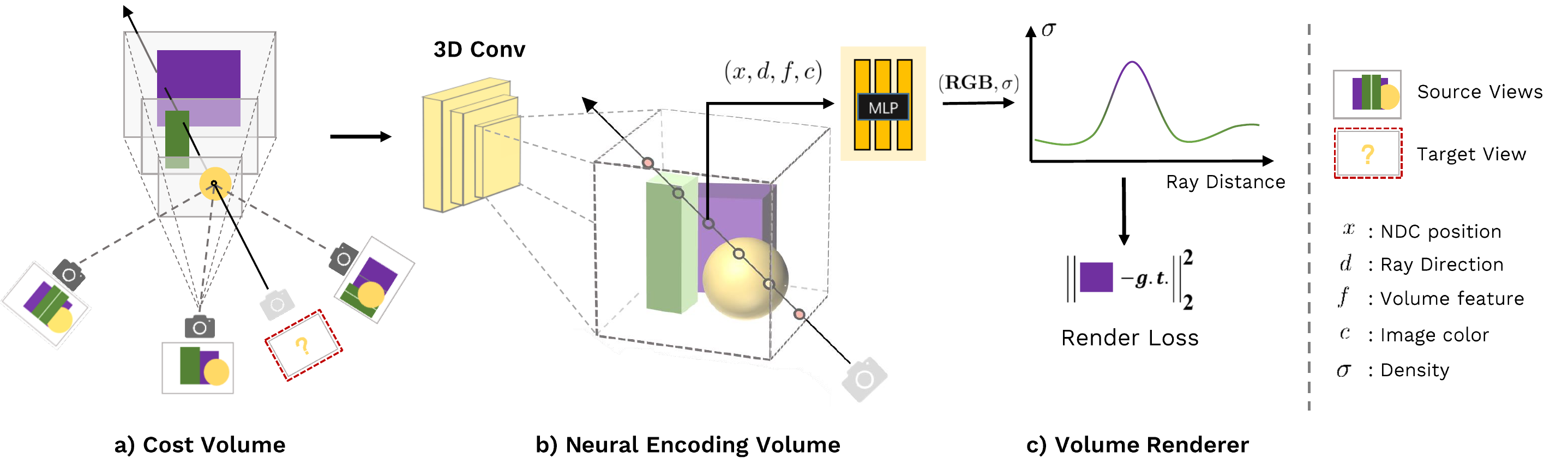}
\end{center}
\caption{Overview of MVSNeRF. Our framework first constructs a cost volume (a) by warping 2D image features onto a plane sweep. We then apply 3D CNN to reconstruct a neural encoding volume with per-voxel neural features (b). We use an MLP to regress volume density and RGB radiance at an arbitrary location using features interpolated from the encoding volume. These volume properties are used by differentiable ray marching for final rendering (c). }
\label{fig:pipeline}
\end{figure*}

We take advantage of the recent success on deep multi-view stereo (MVS) \cite{yao2018mvsnet,gu2020cascade,cheng2020deep}. This line of work can train generalizable neural networks for the task of 3D reconstruction by applying 3D convolutions on cost volumes.
Similar to \cite{yao2018mvsnet}, we build a cost volume at the input reference view by warping 2D image features (inferred by a 2D CNN) from nearby input views onto sweeping planes in the reference view's frustrum. 
Unlike MVS methods \cite{yao2018mvsnet,cheng2020deep} that merely conduct depth inference on such a cost volume, 
our network reasons about both scene geometry and appearance, and outputs a neural radiance field (see Fig.~\ref{fig:pipeline}), enabling view synthesis.
Specifically, leveraging 3D CNN, we reconstruct (from the cost volume) a \emph{neural scene encoding volume} that consists of per-voxel neural features that encode information about the local scene geometry and appearance.
Then, we make use of a multi-layer perceptron (MLP) to decode the volume density and radiance at arbitrary continuous locations using tri-linearly interpolated neural features inside the encoding volume. 
In essence, the encoding volume is a localized neural representation of the radiance field; once estimated, this volume can be used directly (dropping the 3D CNN) for final rendering by differentiable ray marching (as in \cite{nerf}).



Our approach takes the best of the two worlds, learning-based MVS and neural rendering. Compared with existing MVS methods, we enable differentiable neural rendering that allows for training without 3D supervision and inference time optimization for further quality improvement. Compared with existing neural rendering works, our MVS-like architecture is natural to conduct cross-view correspondence reasoning, facilitating the generalization to unseen testing scenes and also leading to better neural scene reconstruction and rendering.
Our approach can, therefore, significantly outperform the recent concurrent generalizable NeRF work \cite{yu2020pixelnerf,ibrnet} that mainly considers 2D image features without explicit geometry-aware 3D structures (See Tab.~\ref{tb:rendering} and Fig.~\ref{fig:rendering}).
We demonstrate that, using only three input images, our network trained from the DTU dataset can synthesize photo-realistic images on testing DTU scenes, and can even generate reasonable results on other datasets that have very different scene distributions.
Moreover, our estimated three-image radiance field (the neural encoding volume) can be further easily optimized on novel testing scenes to improve the neural reconstruction if more images are captured, leading to photo-realistic results that are comparable or even better than a per-scene overfit NeRF, despite of ours taking substantially less optimization time than NeRF (see Fig.~\ref{fig:teaser}). 

These experiments showcase that our technique can be used either as a strong reconstructor that can reconstruct a radiance field for realistic view synthesis when there are only few images captured, or as a strong initializor that significantly facilitates the per-scene radiance field optimization when dense images are available. 
Our approach takes an important step towards making realistic neural rendering practical.  We have released the code at \href{https://apchenstu.github.io/mvsnerf/}{mvsnerf.github.io}.

\section{Related Work}
\boldstart{Multi-view stereo.}
Multi-view stereo (MVS) is a classical computer vision problem, 
aiming to achieve dense geometry reconstruction using images captured from multiple viewpoints, 
and has been extensively explored by various traditional methods \cite{de1999poxels,kutulakos2000theory,kolmogorov2002multi,esteban2004silhouette,seitz2006comparison,furukawa2010accurate,schonberger2016pixelwise}.
Recently, deep learning techniques have been introduced to address MVS problems \cite{yao2018mvsnet,im2018dpsnet}.
MVSNet \cite{yao2018mvsnet} applies a 3D CNN on a plane-swept cost volume at the reference view for depth estimation, achieving high-quality 3D reconstruction that outperforms classical traditional methods \cite{furukawa2010accurate,schonberger2016pixelwise}.
Following works have extended this technique with recurrent plane sweeping \cite{yao2019recurrent}, point-based densification \cite{chen2019point}, confidence-based aggregation \cite{luo2019p}, and multiple cost volumes \cite{cheng2020deep,gu2020cascade}, improving the reconstruction quality.
We propose to combine the cost-volume based deep MVS technique with differentiable volume rendering, enabling efficient reconstruction of radiance fields for neural rendering.
Unlike MVS approaches that use direct depth supervision,
we train our network with image loss only for novel view synthesis. This ensures the network to satisfy multi-view consistency, naturally allowing for high-quality geometry reconstruction.
As a side product, our MVSNeRF can achieve accurate depth reconstruction (despite of no direct depth supervision) comparable to the MVSNet \cite{yao2018mvsnet}.
This can potentially inspire future work on developing unsupervised geometry reconstruction methods.

\boldstartspace{View synthesis.}
View synthesis has been studied for decades with various approaches including light fields \cite{gortler1996lumigraph,levoy1996light,wood2000surface,kalantari2016learning,srinivasan2017learning,chen2018deepsurface}, image-based rendering \cite{debevec1996modeling,buehler2001unstructured,sinha2009piecewise,chaurasia2011silhouette,chaurasia2013depth}, and other recent deep learning based methods \cite{zhou2016view,zhou2018stereo,xu2019deep,flynn2016deepstereo}. 
Plane sweep volumes have also been used for view synthesis \cite{penner2017soft,zhou2018stereo,flynn2016deepstereo,llff,xu2019deep}.
With deep learning, MPI based methods \cite{zhou2018stereo,extremeview,llff,srinivasan2019pushing} build plane sweep volumes at reference views, while other methods \cite{flynn2016deepstereo, xu2019deep} construct plane sweeps at novel viewpoints;
these prior works usually predict colors at the discrete sweeping planes and aggregate per-plane colors using alpha-blending or learned weights.
Instead of direct per-plane color prediction, our approach infers per-voxel neural features in the plane sweep as a scene encoding volume and can regress volume rendering properties from it at arbitrary 3D locations. This models a continuous neural radiance field, allowing for physically based volume rendering to achieve realistic view synthesis.

\boldstartspace{Neural rendering.}
Recently, various neural scene representations have been presented to achieve view synthesis and geometric reconstruction tasks \cite{zhou2018stereo,thies2019deferred,lombardi2019neural,bi2020deep,nerf}.
In particular, NeRF \cite{nerf} combines MLPs with differentiable volume rendering and achieves photo-realistic view synthesis. 
Following works have tried to advance its performance on view synthesis \cite{nerf_in_the_wild,Neural_sparse_voxel_fields};
other relevant works extend it to support other neural rendering tasks like dynamic view synthesis ~\cite{li2020neural,pumarola2021d,sun2021HOI-FVV}, challenge scenes ~\cite{luo2021convolutional,yariv2021volume}, pose estimation ~\cite{meng2021gnerf}, real-time rendering ~\cite{yu2021plenoctrees},  relighting~\cite{rebain2020derf,bi2020neural,chen2020neural}, and editing ~\cite{xiang2021neutex,sofgan}.
We refer the readers to \cite{tewari2020state}  for a comprehensive review of neural rendering.
However, most prior works still follow the original NeRF and require an expensive per-scene optimization process.
We instead leverage deep MVS techniques to achieve across-scene neural radiance field estimation for view synthesis using only few images as input.
Our approach utilizes a plane swept 3D cost volume for geometric-aware scene understanding, achieving significantly better performance than concurrent works \cite{yu2020pixelnerf,ibrnet} that only consider 2D image features for the generalization of radiance field reconstruction. 

\newcommand{\Img}{I}
\newcommand{\Color}{c}
\newcommand{\ImgNum}{M}
\newcommand{\FMap}{F}
\newcommand{\FVec}{f}
\newcommand{\FChNum}{f}
\newcommand{\NetF}{T}
\newcommand{\NetC}{B}
\newcommand{\NetD}{A}
\newcommand{\Cam}{\Phi}
\newcommand{\Pos}{x}
\newcommand{\Dir}{d}
\newcommand{\Rad}{r}
\newcommand{\Hom}{\mathcal{H}}
\newcommand{\Var}{\mathrm{Var}}

\newcommand{\CostV}{P}
\newcommand{\SwpNum}{D}
\newcommand{\SceneV}{S}

\newcommand{\Trans}{\tau}
\newcommand{\Dens}{\sigma}
\newcommand{\Step}{\Delta}

\section{MVSNeRF}
We now present our MVSNeRF. Unlike NeRF ~\cite{nerf} that reconstructs a radiance field via a per-scene "network memorization", our MVSNeRF learns a generic network for radiance field reconstruction.
Given $\ImgNum$ input captured images $\Img_i$ ($i=1,..,\ImgNum$) of a real scene and their known camera parameters $\Cam_i$, we present a novel network that can reconstruct a radiance feild as a neural encoding volume and use it to regress volume rendering properties (density and view-dependent radiance) at arbitrary scene locations for view synthesis. In general, our entire network can be seen as a function of a radiance field, expressed by:
\begin{equation}
\sigma, \Rad = \mathrm{MVSNeRF}(\Pos, \Dir; \Img_i, \Cam_i)
\label{eq:mvsnerf}
\end{equation}
where $\Pos$ represents a 3D location, $\Dir$ is a viewing direction, $\sigma$ is the volume density at $\Pos$, and $\Rad$ is the output radiance (RGB color) at $x$ depending on the viewing direction $\Dir$.
The output volume properties from our network can be directly used to synthesize a novel image $\Img_t$ at a novel target viewpoint $\Cam_t$ via differentiable ray marching.

In this paper, we consider a sparse set of nearby input views for efficient radiance field reconstruction. In practice we use $\ImgNum=3$ views for our experiments, while our approach handles unstructured views and can easily support other numbers of inputs.
The overview of our MVSNeRF is shown in Fig.~\ref{fig:pipeline}. 
We first build a cost volume at the reference view (we refer to the view $i=1$ as the reference view) by warping 2D neural features onto multiple sweeping planes (Sec.~\ref{sec:costvolume}). 
We then leverage a 3D CNN to reconstruct the neural encoding volume, and use an MLP to regress volume rendering properties, expressing a radiance field (Sec.~\ref{sec:radiancefield}).
We leverage differentiable ray marching to regress images at novel viewpoints using the radiance field modeled by our network;
this enables end-to-end training of our entire framework with a rendering loss (Sec.~\ref{sec:training}).
Our framework achieves radiance field reconstruction from few images. 
On the other hand, when dense images are captured, the reconstructed encoding volume and the MLP decoder can also be fast fine-tuned independently to further improve the rendering quality (Sec.~\ref{sec:perscene}).

\subsection{ Cost volume construction.}
\label{sec:costvolume}

Inspired by the recent deep MVS methods \cite{yao2018mvsnet},
we build a cost volume $\CostV$ at the reference view ($i=1$), allowing for geometry-aware scene understanding. 
This is achieved by warping 2D image features from the $m$ input images to a plane sweep volume on the reference view's frustrum.

\boldstartspace{Extracting image features.} 
We use a deep 2D CNN $\NetF$ to extract 2D image features at individual input views to effectively extract 2D neural features that represent local image appearance.
This sub-network consists of downsampling convolutional layers and convert an input image $\Img_i \in \mathbb{R}^{H_i \times W_i \times 3}$ into a 2D feature map $\FMap_i \in \mathbb{R}^{H_i/4 \times W_i/4 \times C}$, 
\begin{equation}
\begin{aligned}
    \FMap_i = \NetF(\Img_i),
\label{eq:extract}
\end{aligned}
\end{equation}
where $H$ and $W$ are the image height and width, and $C$ is the number of resulting feature channels.


\boldstartspace{Warping feature maps.}
Given the camera intrinsic and extrinsic parameters $\Cam=[K, R, t]$, we consider the homographic warping 
\begin{equation}
\begin{aligned}
    \Hom_i(z) = K_i \cdot( R_i \cdot R_1^T + \frac{(t_1-t_i)\cdot n^{T}_1}{z}) \cdot K^{-1}_{1}
\label{eq:homograph}
\end{aligned}
\end{equation}
where $\Hom_i(z)$ is the matrix warping from the view $i$ to the reference view at depth $z$, $K$ is the intrinsic matrix, and $R$ and $t$ are the camera rotation and translation. Each feature map $\FMap_i$ can be warped to the reference view by:
\begin{equation}
\begin{aligned}
    \FMap_{i,z}(u,v) = \FMap_i(\Hom_i(z)[u,v,1]^T),
\label{eq:warping}
\end{aligned}
\end{equation}
where $\FMap_{i,z}$ is the warped feature map at depth $z$, and $(u,v)$ represents a pixel location in the reference view. 
In this work, we parameterize $(u,v,z)$ using the normalized device coordinate (NDC) at the reference view.

\boldstartspace{Cost volume.} The cost volume $\CostV$ is constructed from the warped feature maps on the $\SwpNum$ sweeping planes. We leverage the variance-based metric to compute the cost, which has been widely used in MVS \cite{yao2018mvsnet,cheng2020deep} for geometry reconstruction.
In particular, for each voxel in $\CostV$ centered at $(u,v,z)$, its cost feature vector is computed by:
\begin{equation}
\begin{aligned}
    \CostV(u,v,z) = \Var(\FMap_{i,z}(u,v)),
\label{eq:cost}
\end{aligned}
\end{equation}
where $\Var$ computes the variance across $\ImgNum$ views.

This variance-based cost volume encodes the image appearance variations across different input views; this explains the appearance variations caused by both scene geometry and view-dependent shading effects.
While MVS work uses such a volume only for geometry reconstruction, we demonstrate that it can be used to also infer complete scene appearance and enable realistic neural rendering.

\subsection{Radiance field reconstruction.}
\label{sec:radiancefield}
We propose to use deep neural networks to effectively convert the built cost volume into a reconstruction of radiance field for realistic view synthesis.
We utilize a 3D CNN $\NetC$ to reconstruct a neural encoding volume $\SceneV$ from the cost volume $\CostV$ of raw 2D image feature costs; $\SceneV$ consists of per-voxel features that encode local scene geometry and appearance. An MLP decoder $\NetD$ is used to regress volume rendering properties from this encoding volume. 

\boldstartspace{Neural encoding volume.} 
Previous MVS works \cite{yao2018mvsnet,gu2020cascade,cheng2020deep} usually predict depth probabilities directly from a cost volume, which express scene geometry only.
We aim to achieve high-quality rendering that necessitates inferring more appearance-aware information from the cost volume.
Therefore, we train a deep 3D CNN $\NetC$ to transform the built image-feature cost volume into a new $C$-channel neural feature volume $\SceneV$, where the feature space is learned and discovered by the network itself for the following volume property regression.
This process is expressed by:
\begin{equation}
\begin{aligned}
    \SceneV = \NetC(\CostV).
\label{eq:encoding}
\end{aligned}
\end{equation}
The 3D CNN $\NetC$ is a 3D UNet with downsampling and upsampling convolutional layers and skip connections, which can effectively infer and propagate scene appearance information, leading to a meaningful scene encoding volume $\SceneV$. 
Note that, this encoding volume is predicted in a unsupervised way and inferred in the end-to-end training with volume rendering (see Sec.~\ref{sec:training}). Our network can learn to encode meaningful scene geometry and appearance in the per-voxel neural features;
these features are later continuously interpolated and converted into volume density and view-dependent radiance.

The scene encoding volume is of relative low resolution because of the downsampling of 2D feature extraction; it is challenging to regress high-frequency appearance from this information alone.
We thus also incorporate the original image pixel data for the following volume regression stage, though we later show that this high-frequency can be also recovered in an augmented volume via a fast per-scene fine-tuning optimization (Sec.~\ref{sec:perscene}).

\boldstartspace{Regressing volume properties.}
Given an arbitrary 3D location $\Pos$ and a viewing direction $\Dir$, we use an MLP $\NetD$ to regress the corresponding volume density $\sigma$ and view-dependent radiance $\Rad$ from the neural encoding volume $\SceneV$. As mentioned, we also consider pixel colors $\Color=[\Img(u_i,v_i)]$ from the original images $\Img_i$ as additional input; here $(u_i,v_i)$ is the pixel location when projecting the 3D point $\Pos$ onto view $i$, and $\Color$ concatenates the colors $\Img(u_i, v_i)$ from all views as a $3\ImgNum$-channel vector. The MLP is expressed by:
\begin{equation}
\begin{aligned}
    \sigma, \Rad = \NetD(\Pos, \Dir, \FVec, \Color),\quad \FVec=\SceneV(\Pos),
\label{eq:decoding}
\end{aligned}
\end{equation}
where $\FVec=\SceneV(\Pos)$ is the neural feature trilinearly interpolated from the volume $\SceneV$ at the location $\Pos$.
In particular, $\Pos$ is parameterized in the reference view's NDC space and $\Dir$ is represented by a unit vector at the reference view's coordinate.
Using NDC space can effectively normalize the scene scales across different data sources, contributing to the good generalizability of our method.
In addition, inspired by NeRF \cite{nerf}, we also apply positional encoding on the position and direction vectors ($\Pos$ and $\Dir$), which further enhance the high-frequency details in our results.

\boldstartspace{Radiance field.}
As a result, our entire framework models a neural radiance field, regressing volume density and view-dependent radiance in the scene from few (three) input images. 
In addition, once the scene encoding volume $\SceneV$ is reconstructed, this volume combined with the MLP decoder $\NetD$ can be used independently without the prepending 2D and 3D CNNs. 
They can be seen as a standalone neural representation of the radiance field, outputting volume properties and thus supporting volume rendering.

\begin{figure*}[t]
\begin{center}
    \includegraphics[width=0.95\linewidth]{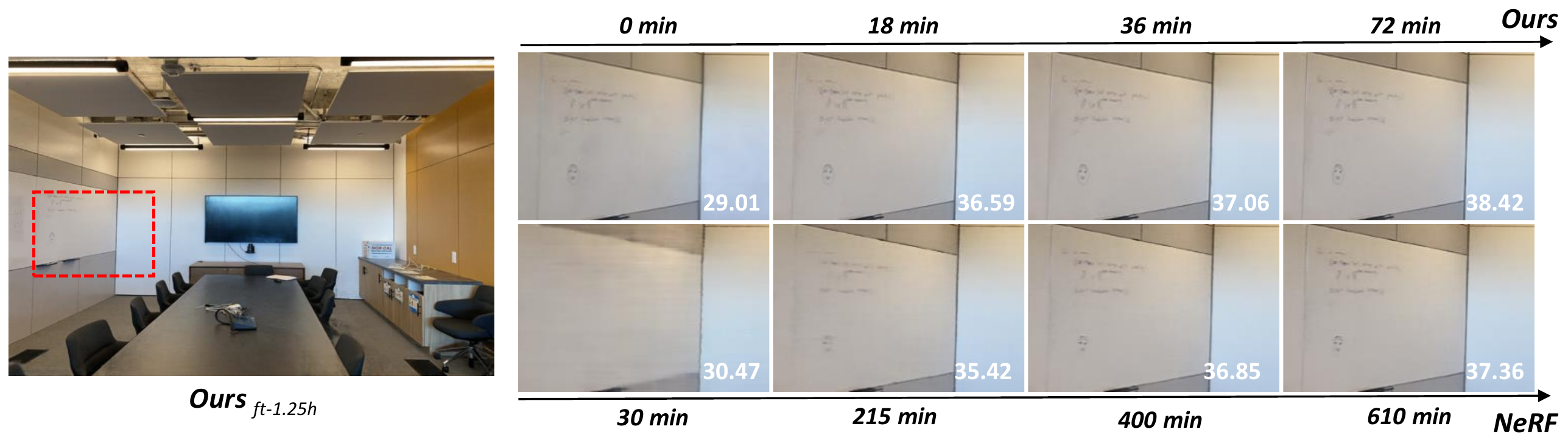}
\end{center}
\caption{Optimization progress. We show results of our fine-tuning (top) and optimizing a NeRF \cite{nerf} (bottom) with different time periods. Our 0-min result refers to the initial output from our network inference. Note that our 18-min results are already much better than the 215-min NeRF results. PSNRs of the image crops are shown in the figure.
}
\label{fig:progress}
\end{figure*}

\subsection{Volume rendering and end-to-end training.}
\label{sec:training}
Our MVSNeRF reconstructs a neural encoding volume and regresses volume density and view-dependent radiance at arbitrary points in a scene. This enables applying differentiable volume rendering to regress images colors. 

\boldstartspace{Volume rendering.}  The physically based volume rendering equation can be numerically evaluated via differentiable ray marching (as is in NeRF \cite{nerf}) for view synthesis. In particular, a pixel's radiance value (color) is computed by marching a ray through the pixel and accumulating radiance at sampled shading points on the ray, given by:
\begin{equation}
\begin{aligned}
    \Color_t &=  \sum_k\Trans_k (1-\exp (-\Dens_k)) \Rad_k, \\
    \Trans_k &= \exp (-\sum_{j=1}^{k-1} \Dens_j ),
\label{eq:raymarching}
\end{aligned}
\end{equation}
where $\Color_t$ is the final pixel color output,  and $\Trans$ represents the volume transmittance. Our MVSNeRF as a radiance field function essentially provides the volume rendering properties $\Dens_k$ and $\Rad_k$ for the ray marching.

\boldstartspace{End-to-end training.} This ray marching rendering is fully differentiable; it thus allows our framework to regress final pixel colors at novel viewpoints using the three input views from end to end.
We supervise our entire framework with the groundtruth pixel colors, using an L2 rendering loss:
\begin{equation}
\begin{aligned}
    L = \|\Color_t - \tilde{\Color}_t \|^2_2,
\label{eq:loss}
\end{aligned}
\end{equation}
where $\tilde{\Color}_t$ is the groundtruth pixel color sampled from the target image $\Img_t$ at a novel viewpoint.
This is the only loss we use to supervise our entire system.
Thanks to the physically based volume rendering and end-to-end training, the rendering supervision can propagate the scene appearance and correspondence information through every network components and regularize them to make sense for final view synthesis. 
Unlike previous NeRF works \cite{nerf,nerf_in_the_wild,Neural_sparse_voxel_fields} that mainly focus on per-scene training,
we train our entire network across different scenes on the DTU dataset.
Our MVSNeRF benefits from the geometric-aware scene reasoning in cost volume processing and can effectively learn a generic function that can reconstruct radiance fields as neural encoding volumes on novel testing scenes enabling high-quality view synthesis.

\subsection{Optimizing the neural encoding volume.}
\label{sec:perscene}
When training across scenes, our MVSNeRF can already learn a powerful generalizable function, reconstructing radiance fields across scenes from only three input images.
However, because of the limited input and the high diversity across different scenes and datasets, it is highly challenging to achieve perfect results on different scenes using such a generic solution. 
On the other hand, NeRF avoids this hard generalization problem by performing per-scene optimization on dense input images;
this leads to photo-realistic results but is extremely expensive.
In contrast, we propose to fine-tune our neural encoding volume -- that can be instantly reconstructed by our network from only few images --  to achieve fast per-scene optimization when dense images are captured.

\boldstartspace{Appending colors.} As mentioned, our neural encoding volume is combined with pixel colors when sent to the MLP decoder (Eqn.~\ref{eq:decoding}). Retaining this design for fine-tuning still works but leads to a reconstruction that always depends on the three inputs. 
We instead achieve an independent neural reconstruction by appending the per-view colors of voxel centers as additional channels to the encoding volume; these colors as features are also trainable in the per-scene optimization.  
This simple appending initially introduces blurriness in the rendering, which however is addressed very quickly in the fine-tuning process.

\boldstartspace{Optimization.}
After appended with colors, the neural encoding volume with the MLP is a decent initial radiance field that can already synthesize reasonable images. We propose to further fine-tune the voxel features along with the MLP decoder to perform fast per-scene optimization when dense images are available.
Note that, we optimize only the encoding volume and the MLP, instead of our entire network. 
This grants more flexibility to the neural optimization to adjust the per-voxel local neural features independently upon optimization; this is an easier task than trying to optimize shared 
convolutional operations across voxels.
In addition, this fine-tuning avoids the expensive network processing of the 2D CNN, plane-sweep warping, and 3D CNN.
As a result, our optimization can therefore be very fast, taking substantially less time than optimizing a NeRF from scratch, as shown in Fig.~\ref{fig:progress}.

Our per-scene optimization leads to a clean neural reconstruction, independent of any input image data (thanks to appending color channels), similar to \cite{nerf,Neural_sparse_voxel_fields}; the dense input images can be therefore dropped after optimization.
In contrast, the concurrent works \cite{yu2020pixelnerf,ibrnet} require retaining the input images for rendering.
Our encoding volume is also similar to Sparse Voxel fields \cite{Neural_sparse_voxel_fields}; however ours is initially predicted by our network via fast inference, instead of the pure per-scene optimization in \cite{Neural_sparse_voxel_fields}.
On the other hand, we can (as future work) potentially subdivide our volume grid in the fine-tuning for better performance as is done in \cite{Neural_sparse_voxel_fields}.
\section{Implementation details}
\label{sec:impl}
\boldstart{Dataset.}
We train our framework on the DTU~\cite{dtu} dataset to learn a generalizable network.
We follow PixelNeRF \cite{yu2020pixelnerf} to partition the data to $88$ training scenes and $16$ testing scenes, and use an image resolution of $512 \times 640$. 
We also test our model (merely trained on DTU) on the Realistic Synthetic NeRF data~\cite{nerf} and the Forward-Facing data \cite{llff}, which have different scene and view distributions from our training set.
For each testing scene, we select $20$ nearby views; we then select 3 center views as input, 13 as additional input for per-scene fine-tuning, and take the remaining 4 as testing views.



\boldstart{Network details.} We use $\FChNum=32$ channels for feature extraction, which is also the number of feature channels in the cost volume and neural encoding volume (before appending color channels). We adopt $\SwpNum=128$ depth hypotheses uniformly sampled from near to far to specify the plane sweep volume. Our MLP decoder is similar to the MLP of NeRF ~\cite{nerf}, but more compact, consisting of $6$ layers. Unlike NeRF reconstructing two (coarse and fine) radiance fields as separate networks, we only reconstruct one single radiance field and can already achieve good results; an extension to coarse-to-fine radiance fields can be potentially achieved at fine-tuning, by optimizing two separate encoding volumes with the same initialization.
For ray marching, we sample $128$ shading points on each marching ray.
We show detailed network structure in the supplementary materials.

We train our network using one \textit{RTX 2080 Ti} GPU. For the across-scene training on DTU, we randomly sample $1024$ pixels from one novel viewpoints as a batch, and use Adam ~\cite{kingma2014adam} optimizer with an initial learning rate of $5e-4$. 



\begin{table*}[ht]
	\centering
	\begin{tabular}{lcccccccc}
		\toprule
		\multirow{2}{*}{Method} & \multirow{2}{*}{Settings} & \multicolumn{3}{c}{Synthetic Data (NeRF ~\cite{nerf_in_the_wild})} & \multicolumn{3}{c}{Real Data (DTU ~\cite{dtu} / Forward-Facing ~\cite{llff})} \\
		
		\cmidrule(lr){3-5}\cmidrule(lr){6-8}
		
		&& \multicolumn{1}{c}{PSNR$\uparrow$} & \multicolumn{1}{c}{SSIM$\uparrow$} & \multicolumn{1}{c}{LPIPS$\downarrow$} &
		\multicolumn{1}{c}{PSNR$\uparrow$} & 
		\multicolumn{1}{c}{SSIM$\uparrow$} & 
		\multicolumn{1}{c}{LPIPS$\downarrow$} \\
		
		\midrule
		 PixelNeRF~\cite{yu2020pixelnerf}  & \multirow{3}{*}{\shortstack{No per-scene \\ optimization}}  & 7.39 & 0.658 & 0.411 &  19.31/11.24 & 0.789/0.486 & 0.382/0.671 \\
		 IBRNet~\cite{ibrnet}  &  & 22.44 & 0.874 & 0.195 &  26.04/21.79 & 0.917/0.786 & 0.190/0.279 \\
		 Ours  &  & \textbf{23.62} & \textbf{0.897} & \textbf{0.176} &  \textbf{26.63}/\textbf{21.93} & \textbf{0.931}/\textbf{0.795} & \textbf{0.168}/\textbf{0.252} \\
		\midrule
		NeRF$_{10.2h}$~\cite{nerf}    &  \multirow{3}{*}{\shortstack{Per-scene \\ optimization}} & \textbf{30.63} & \textbf{0.962} & \textbf{0.093} & 27.01/\textbf{25.97} & 0.902/0.870& 0.263/ 0.236  \\
		IBRNet$_{ft-1.0h}$~\cite{ibrnet}  &  & 25.62  & 0.939 & 0.110 &  \textbf{31.35}/24.88 & \textbf{0.956}/0.861 & \textbf{0.131}/\textbf{0.189} \\
		Ours$_{ft-15min}$  &  & 27.07  & 0.931 & 0.168 &  28.50/25.45 & 0.933/\textbf{0.877} & 0.179/0.192 \\
		\bottomrule
	\end{tabular}
\rule{0pt}{0.01pt}
\caption{\textbf{Quantitative results of novel view synthesis.} We show averaged results of PSNRs, SSIMs and LPISs on three different datasets. 
On the top, we compare our method with concurrent neural rendering methods \cite{yu2020pixelnerf,ibrnet} with direct network inference. On the bottom, we show our fine-tuning results with only 15min optimization ($10k$ iterations), IBRNet 1.0h optimization ($10k$ iterations) and compare with NeRF's \cite{nerf} results with 10.2h optimization ($200k$ iterations).}
\label{tb:rendering}
\end{table*}

\section{Experiments}
We now evaluate our method and show our results.

\begin{figure*}[t]
\begin{center}
    \includegraphics[width=\linewidth, height=0.8\linewidth]{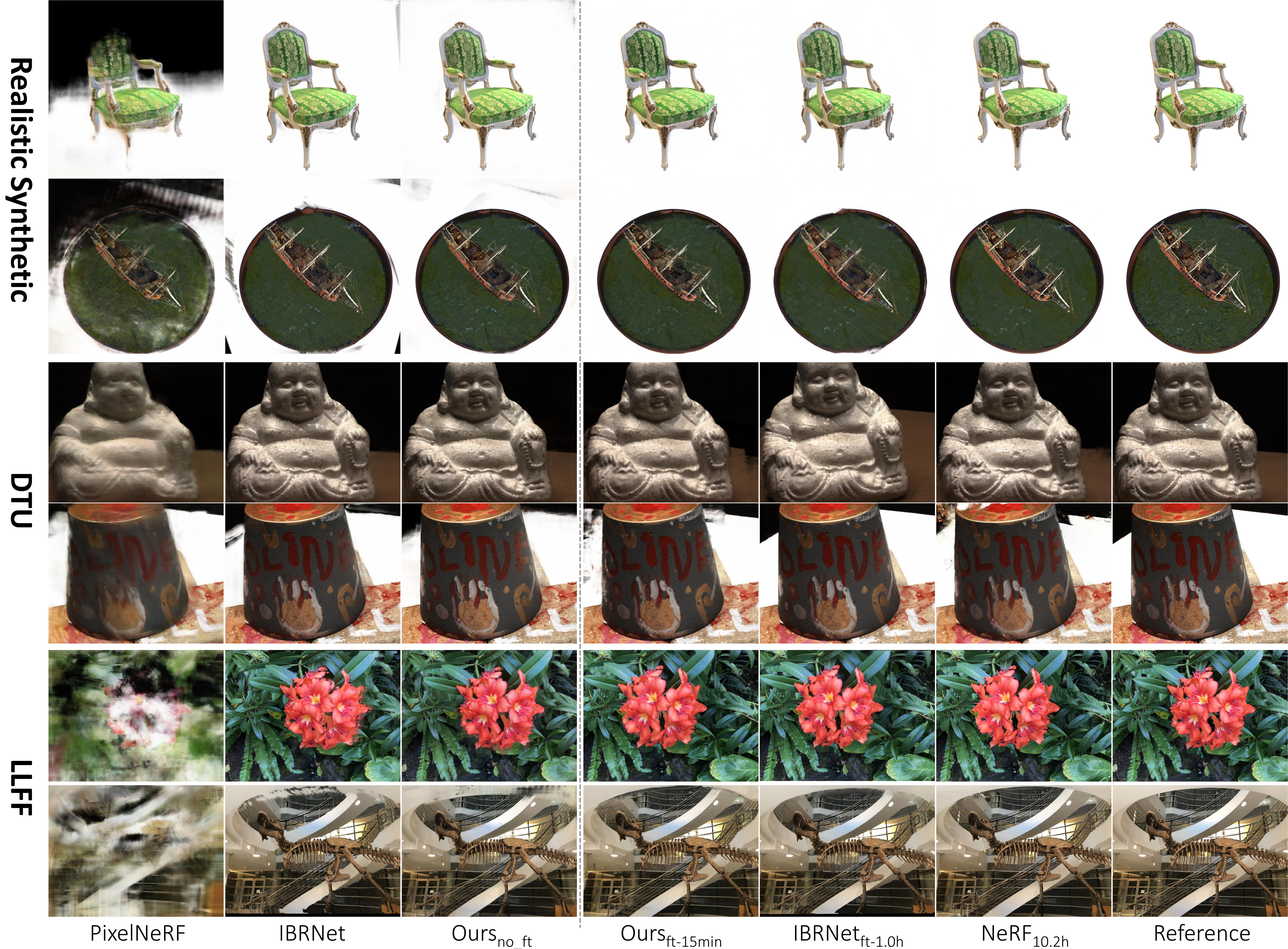}
\end{center}
\caption{Rendering quality comparison. On the left,  we show rendering results of our method and concurrent neural rendering methods \cite{yu2020pixelnerf,ibrnet} by directly running the networks. We show our 15-min fine-tuning results and NeRF's \cite{nerf} 10.2h-optimization results on the right.}
\label{fig:rendering}
\end{figure*}

\boldstartspace{Comparisons on results with three-image input.}
We compare with two recent concurrent works, PixelNeRF\cite{yu2020pixelnerf} and IBRNet \cite{ibrnet} that also aim to achieve the generalization of radiance field reconstruction.
We use the released code and trained model of PixelNeRF and retrain IBRNet on the DTU data (see Sec.~\ref{sec:impl}); we train and test these methods using 3 input views as used in our paper. 
We compare all methods on three datesets \cite{nerf,dtu,llff} with the same input views and use 4 additional images to test each scene.
We show the quantitative results in Tab.~\ref{tb:rendering} and visual comparisons in Fig.~\ref{fig:rendering}.

As shown in Fig.~\ref{fig:rendering}, our approach can achieve realistic view synthesis results using only three images as input across different datasets.
While our model is trained only on DTU, it can generalize well to the other two datesets that have highly different scene and view distributions.
In contrast, PixelNeRF \cite{yu2020pixelnerf} tends to overfit the training setting on DTU.
Although it works reasonably on the DTU testing scenes, it contains obvious artifacts on the Realistic Synthetic scenes and even completely fails on the Forward-Facing scenes.
IBRNet \cite{ibrnet} can do a better job than PixelNeRF when testing on other datasets, but flicker artifacts can still be observed and much more obvious than ours as shown in the appendix video.

These visual results clearly reflect the quantitative results shown in Tab.~\ref{tb:rendering}. 
The three methods can all obtain reasonable PSNRs, SSIMs and LPIPs on the DTU testing set.
However, our approach consistently outperforms PixelNeRF and IBRNet with the same input for all three metrics.
More impressively, our results on the other two testing datasets are significantly better than the comparison methods, clearly demonstrating the good generalizability of our technique.
In general, the two comparison methods both directly aggregate across-view 2D image features at ray marching points for radiance field inference.
Our approach instead leverages MVS techniques for geometry-aware scene reasoning in plane-swept cost volumes, and reconstructs a localized radiance field representation as a neural encoding volume with explicit 3D structures.
This leads to the best generalizablity and the highest rendering quality of our results across different testing scenes.

\begin{table}[t]
	\centering
	\begin{tabular}{lccc}
		\toprule
		Method & \multicolumn{1}{c}{Abs err$\downarrow$} & \multicolumn{1}{c}{Acc (0.01)$\uparrow$} &
		\multicolumn{1}{c}{Acc (0.05)$\uparrow$} \\
		
		\midrule
		MVSNet  & $\textbf{0.018} / \ \ \ - \ \ \  $ & $0.603/ \ \ \ - \ \ \  $ & $\textbf{0.955}/\ \ \  - \ \ \  $ \\
		PixelNeRF  &  $0.245/0.239$ & $0.037/0.039$ &  $0.176/0.187$ \\
		IBRNet &  $ 1.69 / 1.62 $ & $ 0.000 /0.000$ & $ 0.000  /0.001$\\
		Ours &  $0.023/\textbf{0.035}$ & $\textbf{0.746}/\textbf{0.717}$ & $0.913/\textbf{0.866}$\\
		\bottomrule
\end{tabular}
\rule{0pt}{1.0pt}
\caption{\textbf{Depth reconstruction.} We evaluate our unsupervised depth reconstruction on the DTU testing set and compare with other two neural rendering methods (also without depth supervision) PixelNeRF  ~\cite{yu2020pixelnerf} and IBRNet ~\cite{ibrnet}, and a learning based MVS method MVSNet  ~\cite{yao2018mvsnet} that is trained with groundtruth depth. Our method significantly outperforms other neural rendering methods (PixelNeRF and IBRNet) and achieve high depth accuracy comparable to MVSNet. The two numbers of each item refers to the depth at reference/novel views; we mark with "-" when one does not have a reference/novel view. }
\label{tb:geometric}
\end{table}

\boldstartspace{Per-scene fine-tuning results.}
We also show our per-scene optimization results using 16 additional input images in Tab.~\ref{tb:rendering} and Fig.~\ref{fig:rendering}, generated by fine-tuning the neural encoding volume (with the MLP) predicted by our network (Sec.~\ref{sec:perscene}).
Because of the strong initialization obtained from our network, 
we only fine-tune our neural reconstruction for a short period of 15 minutes (10k iterations), which can already lead to photo-realistic results.
We compare our fast fine-tuning results with NeRF's \cite{nerf} results generated with substantially longer optimization time (as long as 10.2 hours).
Note that, our initial rendering results can be significantly boosted with even only 15min fine-tuning; this leads to high-quality results that are on par (Realistic Synthetic) or better (DTU and Forward-Facing) than NeRF's results with 30 times longer optimization time. 
We also show results on one example scene that compare the optimization progresses of our method and NeRF with different optimization times in Fig.~\ref{fig:progress}, which clearly demonstrates the significantly faster convergence of our technique.
By taking our generic network to achieve strong initial radiance field, our approach enables highly practical per-scene radiance field reconstruction when dense images are available.

\boldstartspace{Depth reconstruction.}
Our approach reconstructs a radiance field that represents scene geometry as volume density. We evaluate our geometry reconstruction quality by comparing  depth reconstruction results, generated from the volume density by a weighted sum of the depth values of the sampled points on marched rays (as is done in \cite{nerf}).  We compare our approach with the two comparison radiance field methods \cite{yu2020pixelnerf,ibrnet} and also the classic deep MVS method MVSNet \cite{yao2018mvsnet} on the DTU testing set.
Thanks to our cost-volume based reconstruction, our approach achieves significantly more accurate depth than the other neural rendering methods \cite{yu2020pixelnerf,ibrnet}.
Note that, although our network is trained with only rendering supervision and no depth supervision,  our approach can achieve high reconstruction accuracy comparable to the MVS method \cite{yao2018mvsnet} that has direct depth supervision.
This demonstrates the high quality of our geometry reconstruction, which is one critical factor that leads to our realistic rendering.

\section{Conclusion}
We present a novel generalizable approach for high-quality radiance field reconstruction and realistic neural rendering.
Our approach combines the main advantages of deep MVS and neural rendering,
successfully incorporating cost-volume based scene reasoning into physically based neural volumetric rendering.
Our approach enables high-quality radiance field reconstruction from only three input views and can achieve realistic view synthesis results from the reconstruction.
Our method generalizes well across diverse testing datasets and can significantly outperform concurrent works \cite{yu2020pixelnerf,ibrnet} on generalizable radiance field reconstruction.
Our neural reconstruction can also be fine-tined easily for per-scene optimization, when dense input images are available, allowing us to achieve photo-realistic renderings that are better than NeRF while using substantially less optimization time.
Our work offers practical neural rendering techniques using either few or dense images as input.  

\section{Acknowledgements}
 This work was supported by NSFC programs (61976138, 61977047); the National Key Research and Development Program (2018YFB2100500); STCSM (2015F0203-000-06) and SHMEC (2019-01-07-00-01-E00003); NSF grant IIS-1764078 and gift money from VIVO.









{\small
\bibliographystyle{ieee_fullname}
\bibliography{egbib}
}

\appendix
\newpage

\section{Per-scene optimization.}
\boldstart{More details.} As described in the paper Sec.~3.4, we optimize the predicted neural encoding volume with the MLP decoder for each scene to achieve final high-quality fine-tuning results.
The neural encoding volume with the MLP is an effective neural representation of a radiance field. All fine-tuning results (and also NeRF's optimization comparison results) are generated using a single NVIDIA RTX 2080Ti GPU. On this hardware, our 15min fine-tuning corresponds to 10k training iterations and NeRF's 10.2h optimization corresponds to 200k training iterations.

Our neural encoding volume is reconstructed in the frustrum of the reference view; it thus only covers the scene content in the frustrum. As a result, for a large scene, artifacts may appear when some parts that are not located in the frustrum show up in the novel view. 
Therefore, we extend the neural encoding volume by padding its boundary voxels when fine-tuning on some large scenes. This can address the out-of-frustrum artifacts, though the padding voxels are not well reconstructed initially by the network and may require longer fine-tuning to achieve high quality. 

As described in the paper, we do not apply two-stage (coarse and fine) ray sampling as done in NeRF \cite{nerf}. We uniformly sample points (with per-step small randomness) along each marching ray. We find that 128 points are enough for most scenes and keep using 128 point for our across-scene training on the DTU dataset. When fine-tuning, we increase the number of points to 256 for some challenging scenes. 

\boldstartspace{Optimization progress.}
We have demonstrated in the paper that our 15min fine-tuning results of DTU and LLFF dataset are comparable with the 10.2h optimization results of NeRF \cite{nerf}.
We now show comparisons of the per-scene optimization progress between our fine-tuning and NeRF's from-scratch optimization. 
Note that, thanks to the strong initial reconstruction predicted by our network, our fine-tuning is consistently better than NeRF's optimization through 200k training iterations. As mentioned, our each 18min result corresponds to the result at the 12k-th training iteration, which is at the very early stage in the curves; however, as demonstrated, it can be already better than the NeRF's result after 48k iterations, corresponding to the 10.2h optimization result shown in the paper.
Moreover, while our 15min results are already very good, 
our results can be further improved over more iterations, if continuing optimizing the radiance fields.


\begin{table}[ht]
\begin{tabular}{cccccc}
\toprule\hline
\textbf{Layer}  & \textbf{k} & \textbf{s} & \textbf{d}  & \textbf{chns} & input  \\ \hline
CBR2D$_0$ & 3 & 1 & 1 & $3/8$    & $I$\\
CBR2D$_1$ & 3 & 1 & 1 & $8/8$    & CBR2D$_0$\\
CBR2D$_2$ & 5 & 2 & 2 & $8/16$   & CBR2D$_1$\\
CBR2D$_3$ & 3 & 1 & 1 & $16/16$  & CBR2D$_2$\\
CBR2D$_4$ & 3 & 1 & 1 & $16/16$  & CBR2D$_3$\\
CBR2D$_5$ & 5 & 2 & 2 & $16/32$  & CBR2D$_4$\\
CBR2D$_6$ & 3 & 1 & 1 & $32/32$  & CBR2D$_5$\\
$T$       & 3 & 1 & 1 & $32/32$  & CBR2D$_6$\\ \hline

CBR3D$_0$ & 3 & 1 & 1 & $32+9/8$  & $T, I$\\

CBR3D$_1$ & 3 & 2 & 1 & $8/16$  & CBR3D$_0$\\
CBR3D$_2$ & 3 & 1 & 1 & $16/16$ & CBR3D$_1$\\

CBR3D$_3$ & 3 & 2 & 1 & $16/32$  & CBR3D$_2$\\
CBR3D$_4$ & 3 & 1 & 1 & $32/32$  & CBR3D$_3$\\

CBR3D$_5$ & 3 & 2 & 1 & $32/64$  & CBR3D$_4$\\
CBR3D$_6$ & 3 & 1 & 1 & $64/64$  & CBR3D$_5$\\

CTB3D$_0$ & 3 & 2 & 1 & $64/32$ & CTB3D$_0$ + CBR3D$_4$\\
CTB3D$_1$ & 3 & 2 & 1 & 32/16 & CTB3D$_1$ + CBR3D$_2$\\
CTB3D$_2$ & 3 & 2 & 1 & $16/8$ & CTB3D$_2$ + CBR3D$_0$\\  \hline

PE$_0$     & - & - & - & 3/63 & $x$ \\ 
LR$_0$ & - & - & - & 8+12/256 & $f,c$ \\ 
LR$_1$ & - & - & - & 63/256 & PE\\ 
LR$_{i+1}$ & - & - & - & 256/256 & LR$_i$ $\cdot$ LR$_0$ \\ 
$\sigma$ & - & - & - & 256/1 & LR$_6$\\ 

PE$_1$ & - & - & - & 3/27 & $d$\\ 
LR$_7$ & - & - & - & 27+256/256 & PE$_1$, LR$_6$\\ 
$c$ & - & - & - & 256/3 & LR$_7$\\ 

  \bottomrule
\end{tabular}
\caption{From top to bottom: 2D CNN based feature extraction model, 3D CNN based neural encoding volume prediction model and MLP based volume properties regression model $(i\in[1,...,5])$. \textbf{k} is the kernel size, \textbf{s} is the stride, \textbf{d} is the kernel dilation, and \textbf{chns} shows the number of input and output channels for each layer.  We denote CBR2D/CBR3D/CTB3D/LR to be ConvBnReLU2D, ConvBnReLU3D, ConvTransposeBn3D and LinearRelu layer structure respectively. PE refers to the positional encoding as used in \cite{nerf}. }
\label{tbl:network}
\end{table}

\section{Network Architectures}
We show detailed network architecture specifications of our 2D CNN (that extracts 2D image features ), 3D CNN (that outputs a neural encoding volume), and MLP decoder (that regresses volume properties) in Tab ~\ref{tbl:network}.

\section{Limitations.}
Our approach generally achieves fast radiance field reconstruction for view synthesis on diverse real scenes. However, for highly challenging scenes with high glossiness/specularities, the strong view-dependent shading effects can be hard to directly recovered via network inference and a longer fine-tuning process can be required to fully reconstruct such effects.
Our radiance field representation is reconstructed within the frustrum of the reference view. As a result, only the scene content seen by the reference view is well reconstructed and initialized for the following fine-tuning stage.
Padding the volume (as discussed earlier) can incorporate content out of the original frustrum; however, the unseen parts (including those that are in the frustrucm but are occluded and invisible in the view) are not directly recovered by the network. Therefore, it is challenging to use a single neural encoding volume to achieve rendering in a wide viewing range around a scene (like 360$^\circ$ rendering). Note that, a long per-scene fine-tuning process with dense images covering around the scene can still achieve 360$^\circ$ rendering, though it can be as slow as training a standard NeRF \cite{nerf} (or Sparse Voxel Fields \cite{Neural_sparse_voxel_fields} that is similar to our representation) to recover those uninitialized regions in the encoding volume. Combining multiple neural encoding volumes at multiple views can be an interesting future direction to achieve fast radiance field reconstruction with larger viewing ranges.


\begin{table}[t]
	\centering
	\begin{tabular}{l|ccccc}
        \multicolumn{6}{c}{\textbf{DTU Dataset}}\\ \hline
        Scan &  \#1 & \#8 & \#21 & \#103 & \#114 \\ \hline
		\multicolumn{6}{c}{PSNR$\uparrow$}\\ \hline

		PixelNeRF & 21.64&  23.70& 16.04& 16.76&18.40\\
		IBRNet & 25.97&  \textbf{27.45}& 20.94& 27.91&27.91\\
		Ours & \textbf{26.96} & 27.43 &\textbf{ 21.55}&\textbf{29.25}&\textbf{27.99}\\ \hline
		NeRF$_{10.2h}$ & 26.62& 28.33 & 23.24 & 30.40 &26.47\\
		IBRNet$_{ft-1h}$ & \textbf{31.00}& \textbf{32.46}& \textbf{27.88}& \textbf{34.40}&\textbf{31.00}\\
		Ours$_{ft-15min}$&  28.05 & 28.88 & 24.87 & 32.23 & 28.47 \\ \hline

		\multicolumn{6}{c}{SSIM$\uparrow$}\\ \hline
		PixelNeRF &  0.827& 0.829&  0.691& 0.836&0.763\\
		IBRNet &  0.918& 0.903&  0.873& 0.950&0.943\\
		Ours &  \textbf{0.937} & \textbf{0.922} &\textbf{0.890} &\textbf{0.962}&\textbf{0.949}\\ \hline
		NeRF$_{10.2h}$ &  0.902 & 0.876 &  0.874 & 0.944 & 0.913\\
		IBRNet$_{ft-1h}$ &  \textbf{0.955}& 0.945&  \textbf{0.947}& 0.968&\textbf{0.964}\\
		Ours$_{ft-15min}$& \textbf{0.934} & 0.900& \textbf{0.922}&\textbf{0.964} & 0.945\\\hline

		\multicolumn{6}{c}{LPIPS $\downarrow$ }\\ \hline

		PixelNeRF & 0.373&  0.384& 0.407& 0.376&0.372\\
		IBRNet &  0.190&   0.252& 0.179&  0.195 &  0.136\\
		Ours & \textbf{0.155} & \textbf{0.220} &\textbf{0.166}&\textbf{0.165}&\textbf{0.135}\\ \hline
		NeRF$_{10.2h}$ &  0.265&  0.321&0.246& 0.256 & 0.225\\
		IBRNet$_{ft-1h}$ & \textbf{0.129} &\textbf{0.170}&\textbf{0.104}&\textbf{0.156}&\textbf{0.099} \\
		Ours$_{ft-15min}$ & 0.171 &0.261&0.142&0.170&0.153 \\
		\bottomrule
\end{tabular}
\rule{0pt}{0.05pt}
\caption{\textbf{Quantity comparison on five sample scenes in the DTU testing set}.}
\label{tb:ablation}
\end{table}

\section{Per-scene breakdown.}

We show the pre-scene breakdown analysis of the quantitative results presented in the main paper for the three dataset (\textit{Realistic Synthetic}, \textit{DTU} and \textit{LLFF}).

These results are consistent with the averaged results shown in the paper. In general, since the training set consists of DTU scenes, all three methods can work reasonably well on the DTU testing set.
Our approach can outperform PixelNeRF \cite{yu2020pixelnerf}, when using the same three-image input,  and achieve higher PSNR and SSIM and lower LPIPS. Note that, as mentioned in the paper, the implementation of IBRNet~\cite{ibrnet} is trained and tested with 10 input images to achieve its best performance as used in their paper. Nonetheless, our results with three input images are still quantitatively comparable to the results of IBRNet with 10 input images on the DTU testing set; IBRNet often achieves better PSNRs while we often achieve better SSIMs and LPIPSs.

More importantly, as already shown in paper, when testing on novel datasets, our approach generalizes significantly better than PixelNeRF and IBRNet, leading to much better quantitative results on the Synthetic Data and the Forward-Facing dataset. We also provide detailed per-scene quantitative results for the three testing datasets in Tab.~3-10. Please also refer to the supplementary video for video comparisons.

\begin{table*}[t]
	\centering
	\begin{tabular}{l|cccccccc}
        \hline
        &   Chair & Drums & Ficus & Hotdog & Lego & Materials & Mic & Ship \\ \hline
		\multicolumn{9}{c}{PSNR$\uparrow$  }\\ \hline

		PixelNeRF   &    7.18 & 8.15  & 6.61  & 6.80   & 7.74 & 7.61      & 7.71& 7.30\\
		IBRNet    &    \textbf{24.20} & 18.63 & 21.59  & 27.70  & 22.01 & \textbf{20.91}     & 22.10& 22.36\\
		Ours        &   23.35 & \textbf{20.71} & \textbf{21.98} & \textbf{28.44}  & \textbf{23.18}& 20.05 & \textbf{22.62} & \textbf{23.35}\\ \hline
		NeRF        &    \textbf{31.07}  & \textbf{25.46}  & \textbf{29.73}  & \textbf{34.63} & \textbf{32.66}    &\textbf{30.22}& \textbf{31.81}& 29.49\\
		IBRNet$_{ft-1h}$    &    28.18 & 21.93 & 25.01  & 31.48  & 25.34 & 24.27 & 27.29& 21.48\\
		Ours$_{ft-15min}$ &  26.80 & 22.48 & 26.24 & 32.65 & 26.62 & 25.28 & 29.78 & 26.73 \\\hline

		\multicolumn{9}{c}{SSIM$\uparrow$ }\\\hline

		PixelNeRF & 0.624 & 0.670 & 0.669 & 0.669  & 0.671& 0.644   & 0.729& 0.584\\
		IBRNet   &    0.888& 0.836 & 0.881  & 0.923  & 0.874 & 0.872     & 0.927&  0.794\\
		Ours      &   \textbf{0.876} & \textbf{0.886} & \textbf{0.898} & \textbf{0.962} & \textbf{0.902} & \textbf{0.893} & \textbf{0.923} & \textbf{0.886} \\ \hline
		NeRF      &  \textbf{0.971}& \textbf{0.943}& \textbf{0.969}& \textbf{0.980}&\textbf{0.975}& \textbf{0.968}& \textbf{0.981} &  \textbf{0.908}\\
		IBRNet$_{ft-1h}$    & 0.955 & 0.913 & 0.940  & 0.978  & 0.940 & 0.937 & 0.974& 0.877\\
		Ours$_{ft-15min}$ &  0.934 & 0.898 & 0.944 & 0.971 & 0.924 & 0.927 & 0.970&0.879 \\ \hline

		\multicolumn{9}{c}{LPIPS $\downarrow$}\\ \hline

		PixelNeRF   &  0.386 & 0.421 & 0.335 & 0.433  & 0.427& 0.432     & 0.329 & 0.526\\
		IBRNet   &    \textbf{0.144} &  0.241 & \textbf{0.159} & 0.175 &\textbf{0.202} & \textbf{0.164}  & \textbf{0.103} & 0.369 \\
		Ours        &  0.282& \textbf{0.187} & 0.211 & \textbf{0.173} & 0.204 & 0.216 & 0.177 & \textbf{0.244} \\ \hline
		NeRF        &   \textbf{0.055}&  \textbf{0.101}&\textbf{0.047}& \textbf{0.089}&\textbf{0.054}&0.105& \textbf{0.033}& 0.263\\
		IBRNet$_{ft-1h}$    & 0.079 & 0.133 & 0.082  & 0.093  & 0.105 & \textbf{0.093} & 0.040 & \textbf{0.257}\\
		Ours$_{ft-15min}$ &  0.129& 0.197 & 0.171 & 0.094 & 0.176 & 0.167 & 0.117 & 0.294 \\
		\bottomrule
\end{tabular}
\rule{0pt}{0.05pt}
\caption{\textbf{Quantity comparison on the Realistic Synthetic dataset}.}
\label{tb:ablation}
\end{table*}


\begin{table*}[t]
	\centering
	\begin{tabular}{l|cccccccc}
        \hline
        &  Fern  &Flower&Fortress & Horns & Leaves & Orchids & Room & Trex\\ \hline        
		\multicolumn{9}{c}{PSNR$\uparrow$  }\\ \hline

		PixelNeRF & 12.40 & 10.00 &14.07&11.07& 9.85 &9.62&11.75& 10.55\\
		IBRNet    & 20.83 & 22.38 & \textbf{27.67} & 22.06 & \textbf{18.75} & 15.29 & \textbf{27.26}& 20.06\\
		Ours      & \textbf{21.15} & \textbf{24.74} & 26.03 &\textbf{23.57}&17.51& \textbf{17.85} &26.95&\textbf{23.20}\\\hline
		NeRF$_{10.2h}$ & \textbf{23.87} & 26.84 &\textbf{31.37}&25.96& 21.21 &19.81&\textbf{33.54}& \textbf{25.19}\\
		IBRNet$_{ft-1h}$    & 22.64 & 26.55 & 30.34  & 25.01  & \textbf{22.07} & 19.01 & 31.05 & 22.34\\
		Ours$_{ft-15min}$ &  23.10 & \textbf{27.23}  &30.43 & \textbf{26.35} & 21.54 & \textbf{20.51} & 30.12 & 24.32 \\\hline

		\multicolumn{9}{c}{SSIM$\uparrow$}\\\hline

		PixelNeRF & 0.531 & 0.433 &0.674&0.516& 0.268 & 0.317 &0.691&0.458\\
		IBRNet & \textbf{0.710} & 0.854 &\textbf{0.894}&0.840&\textbf{0.705}&0.571&0.950&  0.768\\
		Ours & 0.638 & \textbf{0.888} &0.872&\textbf{0.868}& 0.667 &\textbf{0.657}&\textbf{0.951} & \textbf{0.868}\\ \hline
		NeRF$_{10.2h}$ & \textbf{0.828}& 0.897 &\textbf{0.945}&0.900& 0.792  &0.721&\textbf{0.978}&\textbf{0.899}\\
		IBRNet$_{ft-1h}$    & 0.774 & 0.909 & 0.937  & 0.904  & \textbf{0.843} & 0.705 & 0.972 & 0.842\\
		Ours$_{ft-15min}$ &  0.795 & \textbf{0.912}&0.943&\textbf{0.917} & 0.826 & \textbf{0.732}&0.966 &0.895\\\hline

		\multicolumn{9}{c}{LPIPS $\downarrow$ }\\\hline

		 &  Fern & Flower & Fortress & Horns & Leaves & Orchids & Room & Trex\\ \hline
		PixelNeRF & 0.650 & 0.708 &0.608&0.705& 0.695 &0.721&0.611& 0.667\\
		IBRNet &0.349 & 0.224 & \textbf{0.196} & 0.285& \textbf{0.292} &0.413&\textbf{0.161}& 0.314\\ 
		Ours   &\textbf{0.238} &  \textbf{0.196}&0.208& \textbf{0.237}& 0.313 &\textbf{0.274}&0.172& \textbf{0.184}\\ \hline
		NeRF$_{10.2h}$ &  0.291& 0.176&0.147& 0.247& 0.301& 0.321 &  0.157 &0.245\\
		IBRNet$_{ft-1h}$    & 0.266 & 0.146 & \textbf{0.133}  & 0.190  & \textbf{0.180} & 0.286 & \textbf{0.089} & 0.222\\
		Ours$_{ft-15min}$ &\textbf{0.253}&\textbf{0.143}&0.134&\textbf{0.188}&0.222&\textbf{0.258}&0.149&\textbf{0.187} \\
		\bottomrule
\end{tabular}
\rule{0pt}{0.05pt}
\caption{\textbf{Quantity comparison on the Forward Facing dataset}.}
\label{tb:ablation}
\end{table*}

\end{document}


\title{Supplementary Material for MVSNeRF}




\maketitle
\ificcvfinal\thispagestyle{empty}\fi

\newcommand{\hao}[1]{{\color{red}[hao: #1]}}



\section{Update the quantitative results on DTU.}
We found an issue in our code when computing the PSNR/SSIM/LPIPS of the results (without fine-tuning) of DTU testing scenes. We therefore show the updated Table 1 of the main paper in Tab.~\ref{tb:rendering}. We apologize for the inconvenience.  Note that, only the DTU numbers of the three methods with no per-scene optimization are
updated. We only consider foreground regions when computing the numbers of DTU results.

In general, since the training set consists of DTU scenes, all three methods can work reasonably well on the DTU testing set.
Our approach can outperform PixelNeRF \cite{yu2020pixelnerf}, when using the same three-image input,  and achieve higher PSNR and SSIM and lower LPIPS. Note that, as mentioned in the paper, the implementation of IBRNet~\cite{ibrnet} is trained and tested with 10 input images to achieve its best performance as used in their paper. Nonetheless, our results with three input images are still quantitatively comparable to the results of IBRNet with 10 input images on the DTU testing set; IBRNet achieves better PSNR while we achieve better SSIM and LPIPS.

More importantly, as already shown in paper, when testing on novel datasets, our approach generalizes significantly better than PixelNeRF and IBRNet, leading to much better quantitative results on the Synthetic Data and the Forward-Facing dataset. We also provide detailed per-scene quantitative results for the three testing datasets in Tab.~3-10. Please also refer to the supplementary video for video comparisons.

\section{Per-scene optimization.}
\boldstart{More details.} As described in the paper Sec.~3.4, we optimize the predicted neural encoding volume with the MLP decoder for each scene to achieve final high-quality fine-tuning results.
The neural encoding volume and the MLP is an effective neural representation of a radiance field. All fine-tuning results (and also NeRF's optimization comparison results) are generated using a single NVIDIA RTX 2080Ti GPU. On this hardware, our 15min fine-tuning corresponds to 10k training iterations and NeRF's 9.5h optimization corresponds to 200k training iterations.

Our neural encoding volume is reconstructed in the frustrum of the reference view; it thus only covers the scene content in the frustrum. As a result, for a large scene, artifacts may appear when some parts that are not located in the frustrum show up in the novel view. 
Therefore, we extend the neural encoding volume by padding its boundary voxels when fine-tuning on some large scenes. This can address the out-of-frustrum artifacts, though the padding voxels are not well reconstructed initially by the network and may require longer fine-tuning to achieve high quality. 

As described in the paper, we do not apply two-stage (coarse and fine) ray sampling as done in NeRF \cite{nerf}. We uniformly sample points (with per-step small randomness) along each marching ray. We find that 128 points are enough for most scenes and keep using 128 point for our across-scene training on the DTU dataset. When fine-tuning, we increase the number of points to 256 for some challenging scenes. 

\begin{figure*}[t]
\begin{center}
    \includegraphics[width=\linewidth]{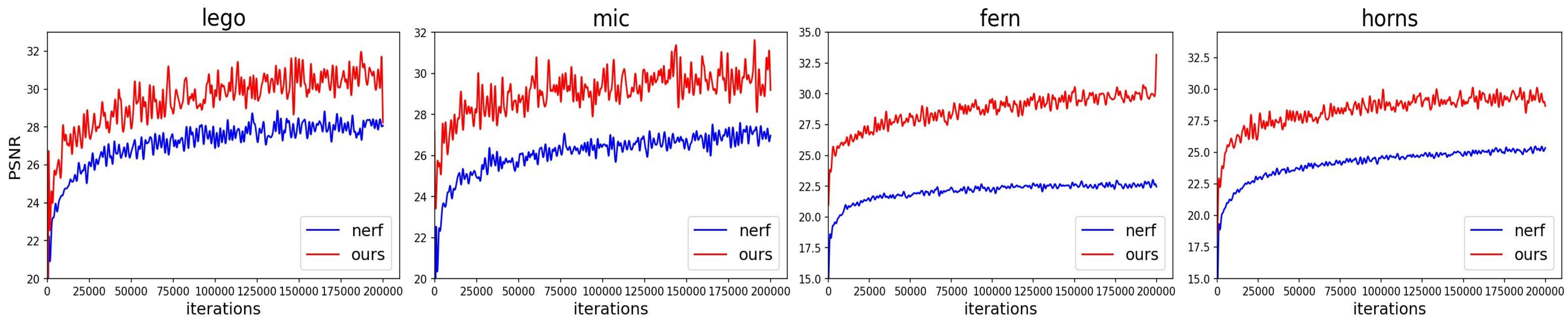}
\end{center}
\caption{Optimization progress. We compare the optimization progress of our fine-tuning with NeRF's optimization from scratch. The PSNRs of 200k training iterations on four testing scenes are plotted. Because of the strong initial radiance field reconstruction from our network, our training is consistently much faster than NeRF through the iterations.}
\label{fig:curves}
\end{figure*}

\boldstartspace{Optimization progress.}
We have demonstrated in the paper that our 15min fine-tuning results can be better than the 9.5h optimization results of NeRF \cite{nerf}.
We now show comparisons of the per-scene optimization progress between our fine-tuning and NeRF's from-scratch optimization. 
The comparisons of the training PSNRs on four testing scenes are shown in Fig~\ref{fig:curves}.
Note that, thanks to the strong initial reconstruction predicted by our network, our fine-tuning is consistently better than NeRF's optimization through 200k training iterations. As mentioned, our each 15min result corresponds to the result at the 10k-th training iteration, which is at the very early stage in the curves; however, as demonstrated, it can be already better than the NeRF's result after 200k iterations, corresponding to the 9.5h optimization result shown in the paper.
Moreover, while our 15min results are already very good, 
our results can be further improved over more iterations, if continuing optimizing the radiance fields, as reflected by the curves.

\begin{table*}[ht]
	\centering
	\begin{tabular}{lcccccccc}
		\toprule
		\multirow{2}{*}{Method} & \multirow{2}{*}{Settings} & \multicolumn{3}{c}{Synthetic Data (NeRF ~\cite{nerf_in_the_wild})} & \multicolumn{3}{c}{Real Data (DTU ~\cite{dtu}, Forward-Facing ~\cite{llff})} \\
		
		\cmidrule(lr){3-5}\cmidrule(lr){6-8}
		
		&& \multicolumn{1}{c}{PSNR$\uparrow$} & \multicolumn{1}{c}{SSIM$\uparrow$} & \multicolumn{1}{c}{LPIPS$\downarrow$} &
		\multicolumn{1}{c}{PSNR$\uparrow$} & 
		\multicolumn{1}{c}{SSIM$\uparrow$} & 
		\multicolumn{1}{c}{LPIPS$\downarrow$} \\
		
		\midrule
		 PixelNeRF~\cite{yu2020pixelnerf}  & \multirow{3}{*}{\shortstack{No per-scene \\ optimization}}  & 4.36 & 0.46 & 0.44 &  24.14/11.266 & 0.887/0.388 & 0.224/0.757 \\
		 IBRNet~\cite{ibrnet}  &  & 19.43 & 0.841 & 0.231 &  \textbf{25.84}/16.70 & 0.902/0.566 & 0.213/0.498 \\
		 Ours  &  & \textbf{22.67} & \textbf{0.90} & \textbf{0.21} &  24.38/\textbf{17.56} & \textbf{0.906}/\textbf{0.691} & \textbf{0.197}/\textbf{0.381} \\
		\midrule
		NeRF$_{9.5h}$~\cite{nerf}    &  \multirow{3}{*}{\shortstack{Per-scene \\ optimization}} & \textbf{26.95} & \textbf{0.939} & \textbf{0.136} & 23.70/\textbf{23.67} & 0.893/0.820& 0.247/ 0.339  \\
		Ours$_{ft-15min}$  &  &  25.17 & 0.919 & 0.257 &  27.72/23.03 & 0.939/0.852 & 0.135/0.210 \\
		Ours$_{ft-75min}$  &  &  25.95 & 0.932 & 0.230 &  \textbf{28.70}/23.37 & \textbf{0.947}/\textbf{0.857} & \textbf{0.116}/\textbf{0.184} \\
		\bottomrule
	\end{tabular}
\rule{0pt}{0.01pt}
\caption{\textbf{Quantitative results of novel view synthesis.} We show averaged results of PSNRs, SSIMs and LPISs on three different datasets. They are a synthesis dataset \cite{nerf} (left) of 8 scenes and two real datasets of 16 DTU (5 of them used for fine tuning evaluation) and 8 Forward-Facing scenes. On the top, we compare our method with concurrent neural rendering methods \cite{yu2020pixelnerf,ibrnet} with direct network inference. On the bottom, we show our fine-tuning results with only 15min and 75min optimization and compare with NeRF's \cite{nerf} results with 9.5h optimization.}
\label{tb:rendering}
\end{table*}

\section{Network Architectures}
We show detailed network architecture specifications of our 2D CNN (that extracts 2D image features ), 3D CNN (that outputs a neural encoding volume), and MLP decoder (that regresses volume properties) in Tab ~\ref{tbl:network}.

\section{Limitations.}
Our approach generally achieves fast radiance field reconstruction for view synthesis on diverse real scenes. However, for highly challenging scenes with high glossiness/specularities, the strong view-dependent shading effects can be hard to directly recovered via network inference and a longer fine-tuning process can be required to fully reconstruct such effects.
Our radiance field representation is reconstructed within the frustrum of the reference view. As a result, only the scene content seen by the reference view is well reconstructed and initialized for the following fine-tuning stage.
Padding the volume (as discussed earlier) can incorporate content out of the original frustrum; however, the unseen parts (including those that are in the frustrucm but are occluded and invisible in the view) are not directly recovered by the network. Therefore, it is challenging to use a single neural encoding volume to achieve rendering in a wide viewing range around a scene (like 360$^\circ$ rendering). Note that, a long per-scene fine-tuning process with dense images covering around the scene can still achieve 360$^\circ$ rendering, though it can be as slow as training a standard NeRF \cite{nerf} (or Sparse Voxel Fields \cite{Neural_sparse_voxel_fields} that is similar to our representation) to recover those uninitialized regions in the encoding volume. Combining multiple neural encoding volumes at multiple views can be an interesting future direction to achieve fast radiance field reconstruction with larger viewing ranges.

\begin{table}[ht]
\begin{tabular}{cccccc}
\toprule\hline
\textbf{Layer}  & \textbf{k} & \textbf{s} & \textbf{d}  & \textbf{chns} & input  \\ \hline
CBR2D$_0$ & 3 & 1 & 1 & $3/8$    & $I$\\
CBR2D$_1$ & 3 & 1 & 1 & $8/8$    & CBR2D$_0$\\
CBR2D$_2$ & 5 & 2 & 2 & $8/16$   & CBR2D$_1$\\
CBR2D$_3$ & 3 & 1 & 1 & $16/16$  & CBR2D$_2$\\
CBR2D$_4$ & 3 & 1 & 1 & $16/16$  & CBR2D$_3$\\
CBR2D$_5$ & 5 & 2 & 2 & $16/32$  & CBR2D$_4$\\
CBR2D$_6$ & 3 & 1 & 1 & $32/32$  & CBR2D$_5$\\
$T$       & 3 & 1 & 1 & $32/32$  & CBR2D$_6$\\ \hline

CBR3D$_0$ & 3 & 1 & 1 & $32+9/8$  & $T, I$\\

CBR3D$_1$ & 3 & 2 & 1 & $8/16$  & CBR3D$_0$\\
CBR3D$_2$ & 3 & 1 & 1 & $16/16$ & CBR3D$_1$\\

CBR3D$_3$ & 3 & 2 & 1 & $16/32$  & CBR3D$_2$\\
CBR3D$_4$ & 3 & 1 & 1 & $32/32$ & CBR3D$_3$\\

CBR3D$_5$ & 3 & 2 & 1 & $32/64$  & CBR3D$_4$\\
CBR3D$_6$ & 3 & 1 & 1 & $64/64$ & CBR3D$_5$\\

CTB3D$_0$ & 3 & 2 & 1 & $64/32$ & CBR3D$_6$\\
CTB3D$_1$ & 3 & 2 & 1 & $64/32$ & CTB3D$_0$ + CBR3D$_4$\\
CTB3D$_2$ & 3 & 2 & 1 & $64/32$ & CTB3D$_1$ + CBR3D$_2$\\
CTB3D$_3$ & 3 & 2 & 1 & $64/32$ & CTB3D$_2$ + CBR3D$_0$\\  \hline

PE$_0$     & - & - & - & 3/63 & $x$ \\ 
LR$_0$ & - & - & - & 8+9/256 & $f,c$ \\ 
LR$_1$ & - & - & - & 63/256 & PE\\ 
LR$_{i+1}$ & - & - & - & 256/256 & LR$_i$+LR$_0$ \\ 
$\sigma$ & - & - & - & 256/1 & LR$_6$\\ 

PE$_1$ & - & - & - & 3/27 & $d$\\ 
LR$_7$ & - & - & - & 27+256/256 & PE$_1$,LR$_6$\\ 
$c$ & - & - & - & 256/3 & LR$_7$\\ 

  \bottomrule
\end{tabular}
\caption{From top to bottom: 2D CNN based feature extraction model, 3D CNN based neural encoding volume prediction model and MLP based volume properties regression model $(i\in[1,...,5])$. \textbf{k} is the kernel size, \textbf{s} is the stride, \textbf{d} is the kernel dilation, and \textbf{chns} shows the number of input and output channels for each layer.  We denote CBR2D/CBR3D/CTB3D/LR to be ConvBnReLU2D, ConvBnReLU3D, ConvTransposeBn3D and LinearRelu layer structure respectively. PE refers to the positional encoding as used in \cite{nerf}. }
\label{tbl:network}
\end{table}

		



\begin{table*}[t]
	\centering
	\begin{tabular}{l|cccccccc}

		\multicolumn{9}{c}{PSNR$\uparrow$ on \textbf{Realistic Synthetic Dataset} }\\

		 &                Chair & Drums & Ficus & Hotdog & Lego & Materials & Mic & Ship \\ \hline
		PixelNeRF   &    3.988 & 5.02  & 4.06  & 4.22   & 4.73 & 4.32      & 4.24& 4.33\\
		IBRNet    &    21.70 & 17.97 & 22.08  & 20.73  & 18.66 & 15.79     & 19.67& 18.85\\
		Ours        &    23.35 & 20.71 & 21.98 & 28.44  & 23.18& 20.05 & 22.62 & 23.35\\ \hline
		NeRF        &    27.21  & 21.91  &25.33  &33.04 & 28.50    &26.38& 26.79& 26.44\\
		Ours$_{ft-75min}$ &  26.60 & 22.35 & 23.60 & 30.94 & 27.1 & 24.26 & 25.84 & 26.62 \\
		\bottomrule
\end{tabular}
\rule{0pt}{0.05pt}
\caption{\textbf{PSNR quantity comparision on NeRF dataset}.}
\label{tb:ablation}
\end{table*}

\begin{table*}[t]
	\centering
	\begin{tabular}{l|cccccccc}

		\multicolumn{9}{c}{SSIM$\uparrow$ on \textbf{Realistic Synthetic Dataset}}\\

		          &  Chair & Drums & Ficus & Hotdog & Lego & Materials & Mic & Ship \\ \hline
		PixelNeRF & 0.428 & 0.485 & 0.479 & 0.492  & 0.477& 0.440   & 0.491& 0.405\\
		IBRNet   &    0.852& 0.817 & 0.895  & 0.863  & 0.823 & 0.285     & 0.899&  0.841\\
		Ours      &   0.876 & 0.886 & 0.898 & 0.962 & 0.902 & 0.893 & 0.923 & 0.886 \\ \hline
		NeRF      &  0.945& 0.901& 0.938& 0.978&0.952& 0.948& 0.959 &  0.892\\
		Ours$_{ft-75min}$ &  0.93 & 0.90 & 0.922 & 0.971 & 0.944 & 0.926 & 0.953&0.904 \\
		\bottomrule
\end{tabular}
\rule{0pt}{0.05pt}
\caption{\textbf{SSIM quantity comparision on NeRF dataset}.}
\label{tb:ablation}
\end{table*}

\begin{table*}[t]
	\centering
	\begin{tabular}{l|cccccccc}

		\multicolumn{9}{c}{LPIPS $\downarrow$ on \textbf{Realistic Synthetic Dataset}}\\

		            &  Chair & Drums & Ficus & Hotdog & Lego & Materials &  Mic  & Ship \\ \hline
		PixelNeRF   &  0.437 & 0.423 & 0.429 & 0.419  & 0.480& 0.414     & 0.433 & 0.461\\
		IBRNet   &    0.214 &  0.255 & 0.142 & 0.238  & 0.251 & 0.285     & 0.174 & 0.285\\
		Ours        &  0.282& 0.187 & 0.211 & 0.173 & 0.204 & 0.216 & 0.177 & 0.244 \\ \hline
		NeRF        &   0.127&  0.188&0.092& 0.111&0.100&0.107& 0.081& 0.282\\
		Ours$_{ft-75min}$ &  0.226 & 0.246 & 0.304 & 0.202 & 0.243 & 0.188 & 0.215 & 0.212 \\
		\bottomrule
\end{tabular}
\rule{0pt}{0.05pt}
\caption{\textbf{LPIPS quantity comparision on NeRF dataset}.}
\label{tb:ablation}
\end{table*}


\begin{table*}[t]
	\centering
	\begin{tabular}{l|cccccccc}

		\multicolumn{9}{c}{PSNR$\uparrow$ on \textbf{LLFF Dataset}}\\

		           &  Fern & Flower &Fortress & Horns & Leaves & Orchids & Room & Trex\\ \hline
		PixelNeRF & 10.00 & 12.23 &13.90&10.90& 10.43 &11.67&11.23& 10.73\\
		IBRNet    & 18.45 & 16.70 &19.94&14.51& 14.84 &12.34& 19.31& 17.51\\
		Ours      & 17.17 & 22.82 & 21.50 &18.73&16.76& 18.90 &15.76&14.12\\\hline
		NeRF$_{9.5h}$ & 22.60 & 25.72 &28.13&23.22& 19.04 &20.25&29.86& 22.58\\
		Ours$_{ft-75min}$ &  22.42 & 27.39  & 21.84 & 27.24 & 18.98 & 19.84 & 27.89 & 25.39 \\
		\bottomrule
\end{tabular}
\rule{0pt}{0.05pt}
\caption{\textbf{PSNR quantity comparision f per-scene optimization on LLFF datasett}.}
\label{tb:ablation}
\end{table*}

\begin{table*}[t]
	\centering
	\begin{tabular}{l|cccccccc}

		\multicolumn{9}{c}{SSIM$\uparrow$ on \textbf{LLFF Dataset}}\\

		  &  Fern & Flower & Fortress & Horns & Leaves & Orchids & Room & Trex\\ \hline
		PixelNeRF & 0.402 & 0.375 &0.590&0.387& 0.192 & 0.223 &0.575&0.350\\
		IBRNet & 0.608 & 0.566 &0.637&0.514&0.444&0.363&0.784&  0.614\\
		Ours & 0.638 & 0.856 &0.861&0.702& 0.668 &0.751&0.687& 0.526\\ \hline
		NeRF$_{9.5h}$ & 0.782& 0.864 &0.921&0.821& 0.676  &0.737&0.966&0.837\\
		Ours$_{ft-75min}$ &  0.797 & 0.926&0.904 & 0.914 & 0.752&0.775 &0.948&0.912\\
		\bottomrule
\end{tabular}
\rule{0pt}{0.05pt}
\caption{\textbf{SSIM quantity comparision of per-scene optimization on LLFF dataset}.}
\label{tb:ablation}
\end{table*}

\begin{table*}[t]
	\centering
	\begin{tabular}{l|cccccccc}

		\multicolumn{9}{c}{LPIPS $\downarrow$ on \textbf{LLFF Dataset}}\\

		 &  Fern & Flower & Fortress & Horns & Leaves & Orchids & Room & Trex\\ \hline
		PixelNeRF & 0.738 & 0.794 &0.725&0.743& 0.754 &0.797&0.782& 0.761\\
		IBRNet &0.453 &  0.498&0.456& 0.576& 0.487 &0.604&0.452& 0.454\\ 
		Ours   &0.407 &  0.212&0.273& 0.377& 0.348 &0.274&0.493& 0.494\\ \hline
		NeRF$_{9.5h}$ &  0.341& 0.265&0.216& 0.400& 0.433& 0.337 &  0.174 &0.339\\
		Ours$_{ft-75min}$ & 0.220&0.107&0.150&0.179&0.239&0.205&0.133&0.159 \\
		\bottomrule
\end{tabular}
\rule{0pt}{0.05pt}
\caption{\textbf{LPIPS quantity comparision  of per-scene optimization on LLFF dataset}.}
\label{tb:ablation}
\end{table*}


\begin{table*}[t]
	\centering
	\begin{tabular}{l|ccccc|ccccc}

		\multicolumn{6}{c}{PSNR$\uparrow$ }& \multicolumn{5}{c}{SSIM $\uparrow$ }\\

		 &  Scan1 & Scan8 & Scan21 & Scan103 & Scan114 &  Scan1 & Scan8 & Scan21 & Scan103 & Scan114 \\ \hline
		PixelNeRF & 23.33&  23.76& 18.77& 22.68&21.03&  0.849& 0.877&  0.729& 0.870&0.763\\
		IBRNet & 26.81&  26.65& 21.23& 27.36&26.06&  0.891& 0.898&  0.811& 0.958&0.902\\
		Ours & 25.33 & 22.96 & 20.92&26.08&18.59 & 0.92 & 0.884 &0.848 &0.931&0.938\\ \hline
		NeRF$_{9.5h}$ & 23.43& 22.35& 22.82& 24.54&25.36&  0.870& 0.892&  0.858& 0.927&0.917\\
		Ours$_{ft-75min}$&  30.61 & 27.05 & 24.58 & 33.06 & 29.07 & 0.959 & 0.934& 0.925&0.972 & 0.958\\
		\bottomrule
\end{tabular}
\rule{0pt}{0.05pt}
\caption{\textbf{PSNR quantity comparision f per-scene optimization on DTU datasett}.}
\label{tb:ablation}
\end{table*}




\begin{table*}[t]
	\centering
	\begin{tabular}{l|ccccc}

		\multicolumn{6}{c}{LPIPS $\downarrow$ on \textbf{DTU Dataset}}\\

		 &  Scan1 & Scan8 & Scan21 & Scan103 & Scan114 \\ \hline
		PixelNeRF & 0.310&  0.301& 0.323& 0.321&0.324\\
		IBRNet &  0.268&   0.268& 0.285&  0.141 &  0.203\\
		Ours & 0.16 & 0.171 &0.206&0.205&0.164\\ \hline
		NeRF$_{9.5h}$ &  0.262&  0.165&0.278& 0.308 & 0.215\\
		Ours$_{ft-75min}$ & 0.109 &0.121&0.130&0.110&0.106 \\
		\bottomrule
\end{tabular}
\rule{0pt}{0.05pt}
\caption{\textbf{LPIPS quantity comparision  of per-scene optimization on DTU dataset}.}
\label{tb:ablation}
\end{table*}

{\small
\bibliographystyle{ieee_fullname}
\bibliography{egbib}
}